\documentclass[lettersize,journal]{IEEEtran}
\usepackage{amsmath,amsfonts}
\usepackage{algorithmic}
\usepackage{algorithm}
\usepackage{array}
\usepackage{multirow}
\usepackage{booktabs}
\usepackage[caption=false,font=normalsize,labelfont=sf,textfont=sf]{subfig}
\usepackage{textcomp}
\usepackage{stfloats}
\usepackage{url}
\usepackage{verbatim}
\usepackage{graphicx}
\usepackage{cite}
\begin{document}

% \title{Predictive Reachability for Learning Mobile Manipulation Behaviors}
\title{Predictive Reachability for Embodiment Selection\\ in Mobile Manipulation Behaviors}

\author{Xiaoxu~Feng, 
        Takato~Horii, and 
        Takayuki~Nagai
% IEEE Publication Technology,~\IEEEmembership{Staff,~IEEE,}
        % <-this % stops a space
\thanks{The authors are with the Graduate School of Engineering Science, Osaka University, Japan (email:x.feng@rlg.sys.es.osaka-u.ac.jp)}}% <-this % stops a space
% \thanks{Manuscript received April 19, 2021; revised August 16, 2021.}}

% The paper headers
\markboth{}%
{Shell \MakeLowercase{\textit{et al.}}: A Sample Article Using IEEEtran.cls for IEEE Journals}

% \IEEEpubid{0000--0000/00\$00.00~\copyright~2021 IEEE}
% Remember, if you use this you must call \IEEEpubidadjcol in the second
% column for its text to clear the IEEEpubid mark.

\maketitle
\pagestyle{empty}
\thispagestyle{empty}

\begin{abstract}
Mobile manipulators require coordinated control between navigation and manipulation to accomplish tasks. 
Typically, coordinated mobile manipulation behaviors have base navigation to approach the goal followed by arm manipulation to reach the desired pose. 
Selecting the embodiment between the base and arm can be determined based on reachability.
Previous methods evaluate reachability by computing inverse kinematics and activate arm motions once solutions are identified. 
In this study, we introduce a new approach called predictive reachability that decides reachability based on predicted arm motions. 
Our model utilizes a hierarchical policy framework built upon a world model. 
The world model allows the prediction of future trajectories and the evaluation of reachability. The hierarchical policy selects the embodiment based on the predicted reachability and plans accordingly.
Unlike methods that require prior knowledge about robots and environments for inverse kinematics, our method only relies on image-based observations. 
We evaluate our approach through basic reaching tasks across various environments. The results demonstrate that our method outperforms previous model-based approaches in both sample efficiency and performance, while enabling more reasonable embodiment selection based on predictive reachability.
\end{abstract}


\section{Introduction}
\IEEEPARstart{M}{obile} manipulation (MM) robots perform tasks over large areas by coordinating base navigation and arm manipulation.
When executing tasks such as reaching and grasping, these robots must approach the target using base movements before activating arm movements to complete tasks. 
Proper selection between base and arm is crucial to prevent safety issues due to arm motions during base navigation.
A focus is to determine appropriate base poses along a trajectory where the arm should be activated.
A straightforward approach involves assessing the reachability with respect to the arm.
The rule dictates that base motion is preferred when the goal is outside the arm’s reachable region, while arm motion is executed if the reachability is identified.
For a base-fixed manipulator, reachability refers to the existence of joint configurations that allow the arm to reach specified poses. 
For mobile manipulators, reachability has to be evaluated at each base pose along a trajectory, and the arm is activated once desired joint configurations are found.

This study explores methods for determining arm reachability to enable effective embodiment selection in coordinated mobile manipulation behaviors.
Two primary methods have been utilized for reachability checks. 
The first is online computation, which solves inverse kinematics (IK) for the arm at each base pose of a trajectory \cite{honerkamp2021learning, jauhri2022robot}. 
The second method involves offline computation, which pre-constructs a representation of the arm’s capabilities to enable quick reachability queries during online processing \cite{zacharias2007capturing}. 
However, both methods require prior knowledge of robots and environments for IK computation, which may limit their applicability in complex environments where such knowledge is challenging to obtain.

\begin{figure}[!t]
\centering
\includegraphics[width=\linewidth]{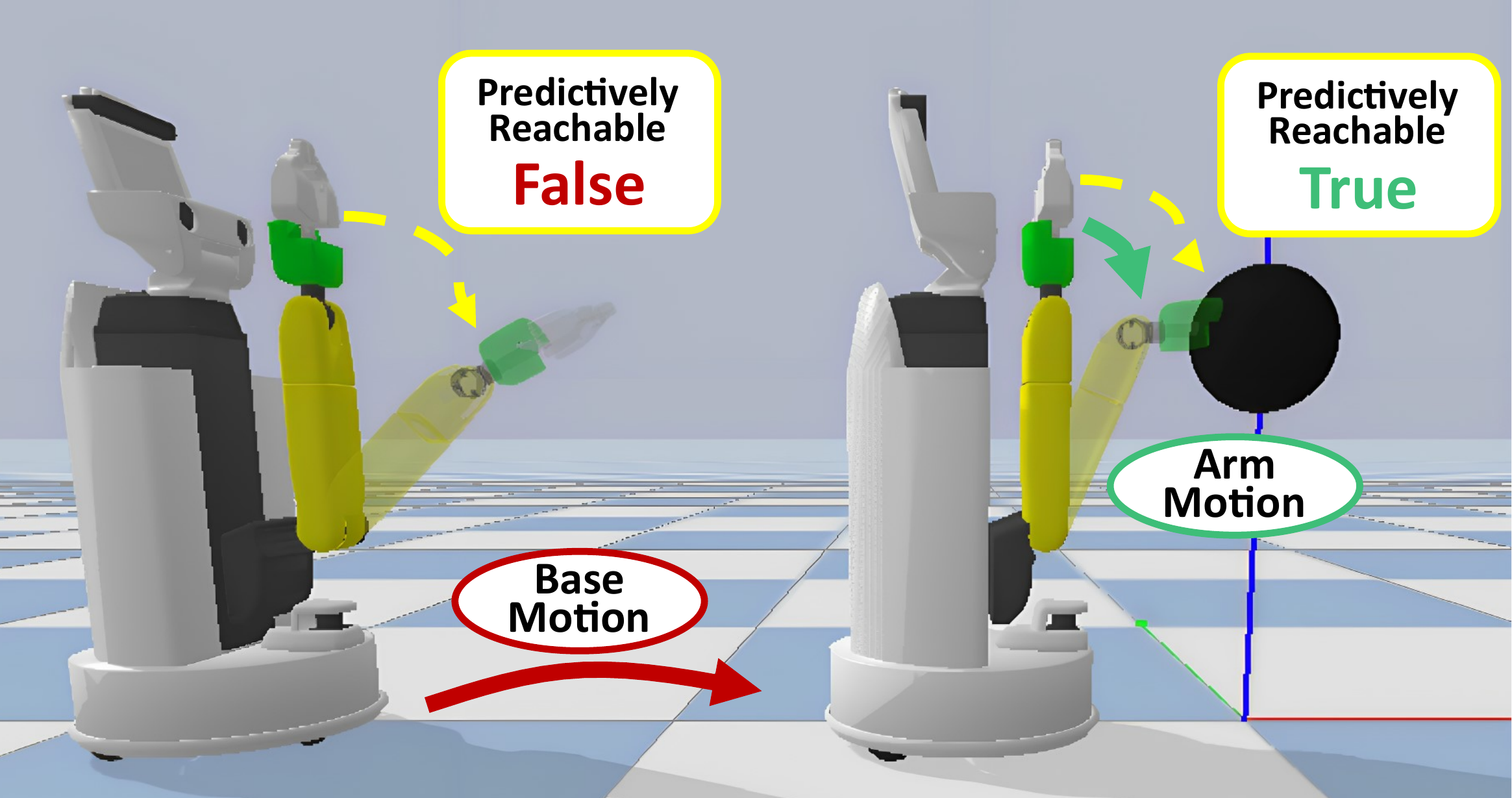}
\caption{Embodiment selection based on predictive reachability. It prefers the arm when goal-reaching is predicted within imagined rollouts. Otherwise, the base is selected to move.}
\label{fig:predictive reachability}
\end{figure}

Vision-based approaches have gained significant traction in robotics due to their rich perceptual capabilities and adaptability. 
We argue that integrating reachability assessment into vision-based systems can significantly broaden the scope of mobile manipulation applications. 
In this work, we propose predictive reachability, which determines reachability based on the arm states in predicted trajectories.
Rather than relying on traditional IK solutions, we employ a learned dynamic model to forecast a future trajectory from an initial state, using actions generated by a policy.
A reward model then serves as a goal-reaching predictor, evaluating whether the arm states in the predicted trajectory can achieve the desired goal.
Recent advancements in world model \cite{ha2018recurrent, hafner2019planet, hafner2019dream, hafner2020dreamerv2, hafner2022deep, hafner2023dreamerv3} provide templates for integrating dynamic models, reward models, and policies into a vision-based system. 
Building on this, we extend existing models to predict reachability based on visual information.

\begin{figure}[!t]
\centering
\includegraphics[width=\linewidth]{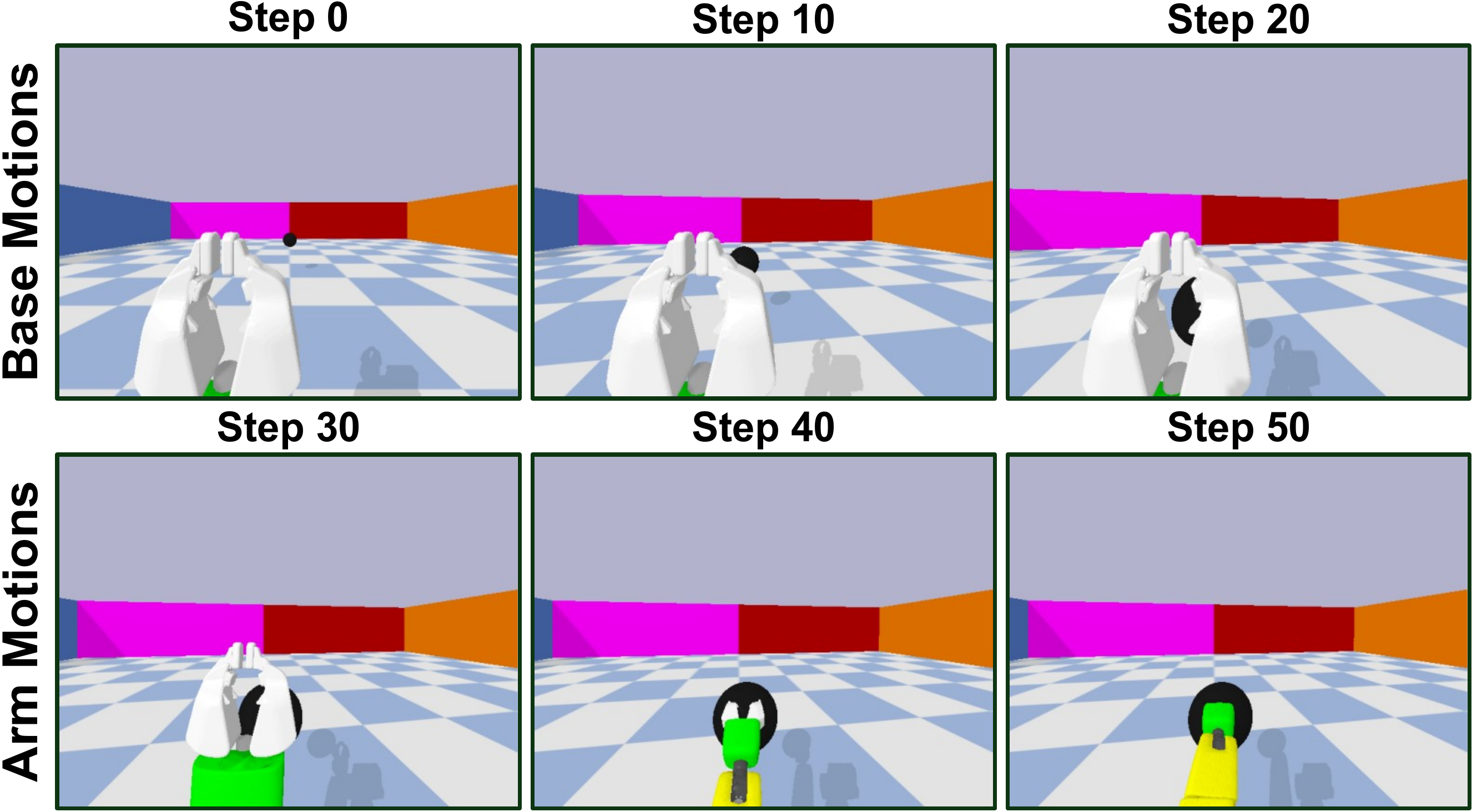}
\caption{Demonstrations of coordinated mobile manipulation behaviors. Each has two stages where arm motions are activated after the base has approached the goal.}
\vspace{-0.5cm}
\label{fig:behavior}
\end{figure}

We introduce Mobile Manipulation Director (MMDirector), which applies predictive reachability for embodiment selection and plans accordingly for coordinated mobile manipulation behaviors, as in Fig.\ref{fig:predictive reachability}. 
MMDirector integrates a world model that compresses image observations into latent states, facilitating the learning of dynamic and reward models within that latent space.
Additionally, MMDirector features a three-level hierarchical policy that takes these latent states as inputs: the highest-level policy manages embodiment selection, while two lower-level policies generate subgoals and actions for embodiment-conditioned planning.
Specifically, the highest policy is trained using a novel reward function based on predictive reachability. 
We incorporate demonstrations as in Fig.\ref{fig:behavior} into the replay buffer and goal-conditioned policy to address the exploration challenges when planning with image-based observations and sparse task rewards.
We evaluate our model on reaching tasks in different environments and provide results showing its effectiveness. 
Additionally, we visualize base placements where the arm is selected to demonstrate the role of predictive reachability.

In summary, the main contributions of this study have:
\begin{enumerate}
    \item We propose predictive reachability, in which learned dynamic and reward models are employed to forecast future reachability given a starting state.
    \item We introduce a reachability reward that encourages the policy to select the arm when predicted reachability is positive and to select the base otherwise.
    \item We propose MMDirector, which incorporates a vision-based world model and a hierarchical policy to achieve embodiment selection and corresponding planning.
\end{enumerate}

\section{Related Work}

\subsection{Mobile Manipulation}
The domain of motion planning for MM has sought to tackle the intricate challenge of coupling navigation and manipulation \cite{siciliano2016handbook}. 
This coupling allows MM robots to manage all active components uniformly.
Toyota develops a whole-body hybrid IK solver to explore feasible trajectories \cite{yamamoto2019development}. 
Reinforcement learning (RL) has garnered attention for learning whole-body MM behaviors \cite{wang2020learning,kindle2020whole}, as it can be easily adapted from other robotic scenarios to MM.
On the other hand, separating navigation and manipulation facilitates more flexible coordination.
In \cite{reister2022combining}, navigation and manipulation costs can be combined to find consecutive and efficient robot placements.
IK is commonly used to compute feasible arm trajectories, and the arm reachability guides the planning of base motion \cite{honerkamp2021learning, jauhri2022robot}. 
HRL4IN \cite{li2020hrl4in} and ReLMoGen \cite{xia2021relmogen} employ a hierarchy that performs embodiment selection at the high level and generates embodiment-conditioned actions at the low level. 
Our work leverages model-based RL to execute embodiment selection and separated planning.

\subsection{Reachability}

In MM behaviors, consecutively assessing reachability for desired poses during base movements is crucial.
Two primary methods are used for the assessment.
The first method is online computation, which repeatedly solves IK at all base positions along trajectories.
During base movements, desired poses for the end-effector can be updated using midpoints between the current and final poses \cite{honerkamp2021learning}.
The reachability at each step can serve as a reward function for training the base policy.
Alternatively, the desired pose can consistently be the final target \cite{jauhri2022robot}.
In this case, arm motion is executed to reach the goal only when reachability is confirmed, and distance-based and task-specific rewards are necessary for training the base policy.
The second method is offline computation, which pre-constructs a representation of the arm's capabilities \cite{zacharias2007capturing}, enabling quick reachability queries during online processing.
Inverse reachability maps (IRMs) have been proposed for MM tasks to represent base placements given a target pose.
IRMs can be learned from datasets \cite{stulp2012ARPlaces} or obtained by the inversion of reachability maps \cite{vahrenkamp2013robot, burget2015stance, makhal2018reuleaux}.
In this work, We leverage the dynamic model and reward predictor in the world model of Director \cite{hafner2022deep} to predict future arm trajectories and identify reachability within them.
Since the world model takes images as input, our method can evaluate predictive reachability in vision-based scenarios, without the need for prior knowledge about robots and environments in IK computations.

\subsection{Hierarchical RL}

Hierarchical RL (HRL) provides a framework where high-level decisions set conditions for lower-level policies \cite{pateria2021hierarchical}.
Typically, the high-level policy generates subgoals, while the low-level policy produces actions to achieve these subgoals, effectively managing long-horizon tasks \cite{hafner2022deep}.
In MM behaviors, embodiment selection influences planning by separating base and arm operations.
In HRL4IN \cite{li2020hrl4in}, the high-level policy predicts both a subgoal and an embodiment selector, while the low-level policy generates actions corresponding to the selected embodiment to achieve the subgoal.
ReLMoGen \cite{xia2021relmogen} follows a similar hierarchy that utilizes motion generators to produce actions.
Our model draws inspiration from the embodiment selection in HRL4IN but introduces a novel three-level policy architecture.
The highest level selects the embodiment, which then conditions the generation of subgoals and actions at the two lower levels.

\section{Method}
\begin{figure*}[!t]
\centering
\includegraphics[width=\linewidth]{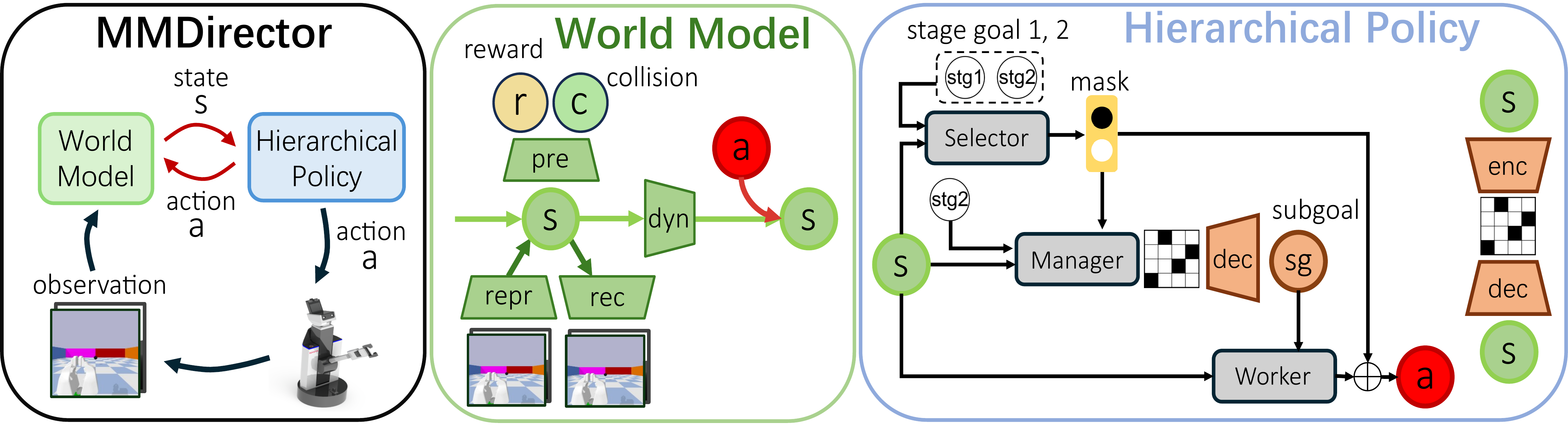}
\caption{Overview of MMDirector. MMDirector comprises a world model and a three-level hierarchical policy. The world model compresses observations into latent states and predicts ahead in that latent space. It is trained through observation reconstruction and learns to predict task rewards and collisions, which are used for reachability evaluations. The selector takes the current state and two stage goals from demonstrations as input to determine embodiment selection. The manager, conditioned on the final goal and embodiment selection, produces discrete codes that are decoded into subgoals within the latent space. The worker generates actions conditioned on subgoals, which are masked by the embodiment selection. All three levels of the policy are trained concurrently using actor-critic algorithms to maximize their respective reward functions.}
\label{fig:Model}
\vspace{-0.3cm}
\end{figure*}

We address the problem of planning MM reaching behaviors with embodiment selection. We formulate this as a model-based RL problem and show that embodiment selection can be achieved based on predictive reachability. Our proposed MMDirector comprises two main components: a world model and a three-level hierarchical policy. The world model learns a dynamic model and a task reward predictor. In a reaching task, a positive task reward indicates successful goal-reaching. Therefore, future rollouts can be predicted using the dynamic model, and reachability within these rollouts can be evaluated using the task reward predictor. The three-level hierarchical policy produces masks for embodiment selection and generates subgoals and actions for planning. The world model is trained using replay buffer experience, and the policy is optimized through actor-critic algorithms. Specifically, the policy for embodiment selection learns from rewards based on predictive reachability, while the others use sparse task rewards and dense rewards computed from the predicted rollouts. Additionally, we incorporate demonstrations into the replay buffer and goal-conditioned policies. The overview is shown in Fig.\ref{fig:Model}.

\begin{figure}[!t]
\centering
\includegraphics[width=\linewidth]{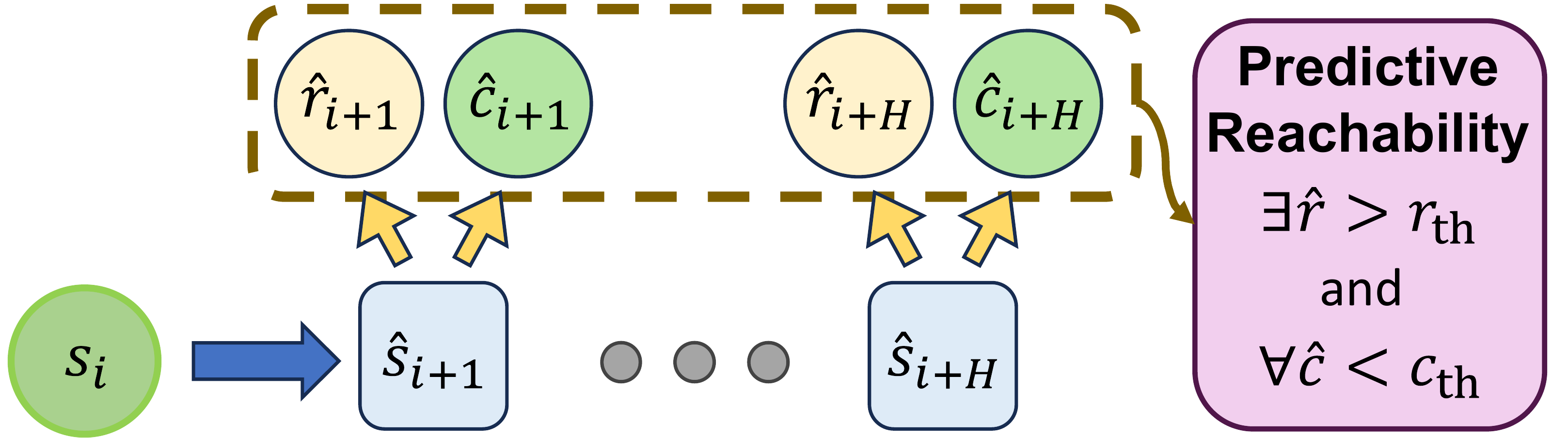}
\caption{Evaluation of predictive reachability. From one latent state, a rollout is predicted. Task rewards and collisions are then predicted given the rollout. Predictive reachability is identified if the sequence of task rewards and collisions satisfies the conditions.}
% \vspace{-0.5cm}
\label{fig:PRI}
\end{figure}

\subsection{Demonstration Collection}
We collect demonstrations of MM reaching behaviors to provide successful experiences that facilitate the learning process. Using the guided policy search algorithm \cite{levine2013guided,levine2016end}, we collect whole-body behaviors and modify them to include separated base and arm stages, as shown in Fig.\ref{fig:behavior}. This ensures that the arm is activated only when reachability is confirmed. Demonstrations are loaded in the buffer. The last steps of the two stages, referred to as stage goals, serve as goals of the policies, and the goal of the arm stage is also known as the final goal. We record image observations including RGB and depth images, proprioceptive states, and actions. We also label selection masks, subgoals, stage goals, task rewards, and collision signals.

\subsection{Predictive reachability and reachability reward}

\begin{table}[!t]
  \centering
  \caption{Reachability reward assignment}
  \label{tab:reac}  % Add a label for referencing
    \begin{tabular}{c|cc}
         \toprule
         \multirow{2}{*}{Selected Embodiment} &
            \multicolumn{2}{c}{Predictive Reachability} \\
         \cmidrule{2-3}
              & True & False \\
         \midrule
         Base & 0 & 0 \\
         Arm  & 2 & -1 \\
         \bottomrule
\end{tabular}
\vspace{-0.3cm}
\end{table}

The world model can learn latent representations and build a dynamic model in the latent space. Given the representation at an initial step $s_t$, the dynamic model forecasts future states $\{s_t, \hat{s}_{t+1:t+H}\}$ using actions $\{a_{t:t+H-1}\}$ generated by the policy, as shown in Fig.\ref{fig:PRI}. $H=18$ is the prediction horizon. Embodiment is selected at the initial step $s_t$ and remains constant throughout the rollout. Task rewards $r$ and collision signals $c$ are predicted based on the future rollout. We define that predictive reachability at the initial state $s_t$ is identified, if, during the rollout, at least one task reward exceeds a threshold $r_{th}=0.7$, and all collision signals are below a threshold $c_{th}=0.3$. 
In each training iteration, one rollout is predicted for each state and the $\epsilon$-greedy algorithm is used in the action generation for stochasticity and exploration. 
Predictive reachability within the rollout is evaluated independently for each state.

We propose a reachability reward to train the policy for embodiment selection, as shown in Table \ref{tab:reac}. This reward depends on the predicted reachability and the chosen embodiment. A positive reward is given when predictive reachability is confirmed and the arm is selected. This aligns with the desired MM behavior that activates the arm only when it can feasibly reach the goal. By contrast, selecting the arm when reachability is false results in a negative reward. 0 is assigned when the base is selected, as the determination of arm reachability is irrelevant to base motions.

\subsection{Model}
Our proposed MMDirector contains a world model base and a three-level hierarchical policy. The world model compresses observations into latent representations and learns a dynamic model and predictors for task rewards and collisions within the latent space. The hierarchical policy takes latent representations as input and produces masks for embodiment selection and generates subgoals and actions for planning.

\subsubsection{World Model}
The world model uses RSSM in PlaNet \cite{hafner2019planet}, including the following components:
\begin{equation}
    \begin{aligned}
    & \text{Representation model:} & &\mathrm{repr_\theta}(s_t|s_{t-1},a_{t-1},x_t) \\
    & \text{Dynamic model:} & &\mathrm{dyn_\theta}(s_t|s_{t-1},a_{t-1}) \\
    & \text{Decoder model:} & &\mathrm{rec_\theta}(x_t|s_t) \\ 
    & \text{Predictor model:} & &\mathrm{pre_\theta}(g_t|s_t)
    \end{aligned}
\end{equation}
The representation model encodes observations $x_t$ with actions $a_{t-1}$ into latent states $s_t$ and the dynamic model predicts future latent states. The decoder re-obtains observations to compute reconstruction loss. 
Predictor models estimate the task reward $r_t$ and collision signal $c_t$, so $g \in \{r, c\}$. 

The world model is optimized end-to-end on sequences, sampled from demonstrations and online samples, on the variational objective:
\begin{multline}
    \mathcal{L}(\theta) \doteq \sum_{t=1}^T \Bigl( \sum_{g \in \{r, c\}}(\mathrm{pre_\theta}(g_t|s_t)-g_t)^2 + \|\mathrm{rec_\theta}(x_t|s_t)-x_t\|^2 \\
    + \beta\mathrm{KL}[\mathrm{repr_\theta}(s_t|s_{t-1},a_{t-1},x_t) \| \mathrm{dyn_\theta}(s_t|s_{t-1},a_{t-1})]\Bigr)
\end{multline}
Latent states obtained from demonstrations and online samples serve as initial steps for predicting latent rollouts. Additionally, we collect two sets of representations of stage goals from demonstrations and compute their means, denoted as $\{stg_1, stg_2\}$, which are used in goal-conditioned policies.

\subsubsection{Top-level selector policy}

The top-level selector policy generates a two-element one-hot mask to select the embodiment. The mask is updated at every step during simulated executions and only once at the initial state for imagined rollouts. We apply a greedy algorithm to the mask to explore embodiments sufficiently. The mask conditions subgoal generation and filters actions to ensure only actions corresponding to the selected embodiment are generated. The selector's input includes the current state and two stage goals. Formally, the selector policy can be represented as 
\begin{equation}
    \begin{aligned}
        \text{Selector Policy:} & & \mathrm{str_\zeta}(m_t|s_t,stg_1,stg_2)
    \end{aligned}
\end{equation}

The selector policy learns a state-action-value critic based on the reachability reward. Since predictive reachability relies only on the current state, we employ a weighted update with a dynamic coefficient to compute the target value of the critic for stability. The loss for the critic and dynamic coefficient are denoted as 
\begin{equation}
    \begin{aligned}
        \mathcal{L}(q) &= \mathrm{E}[\tfrac{1}{2}(q(s_t,a_t)-\mathrm{sg}(Q))^2]\\
        Q &= \mu r+(1-\mu)Q \\
        \mu &= \mu_{\mathrm{predefined}}-10*l_{rew}
    \end{aligned}
\end{equation}

The coefficient $\mu$ is calculated empirically based on the prediction loss of task reward $l_{rew}$ of demonstrations during world model optimization. This helps mitigate the impact of prediction errors of the task reward. $\mathrm{sg}$ refers to the stop gradient operator. The actor’s optimization objective combines policy gradient and policy entropy, denoted as 
\begin{equation}
    \mathcal{L}(\pi) = -\mathrm{E}[\ln\pi(a_t|s_t)\mathrm{sg}(q(s_t,\pi(s_t)))+\eta\mathrm{H}[\pi(a_t|s_t)]]
\end{equation}

\subsubsection{Mid-level manager policy}

The mid-level manager policy generates latent subgoals $sg$ for the lower worker policy to achieve. Subgoals are updated every $K=6$ steps in imagined rollouts and every step during simulated executions. The manager policy's input includes current states, the final goal, and the selection mask. We continue applying the goal autoencoder in Director \cite{hafner2022deep} to prevent directly generating subgoals in the high-dimensional latent space. The autoencoder is trained independently of policy training. Consequently, the manager outputs discrete codes that are then decoded to high-dimensional subgoals by the decoder:

\begin{equation}
    \begin{aligned}
       & \text{Manager Policy:} & & \mathrm{mgr_\phi}(z|s_t,stg_2,m_t) \\
       & \text{Subgoal Encoder:} & & \mathrm{enc}(z|s_t) \\
       & \text{Subgoal Decoder:} & & \mathrm{dec}(s_t|z) \\
    \end{aligned}
\end{equation}

Both manager and worker policies are trained from the imagined rollouts. The manager learns three state-value critics for task rewards, collision signals, and progress rewards. Progress rewards are calculated as the sum of similarities between states of the next $K$ steps and the final goal. This dense reward effectively guides the manager policy toward the goal. Exploration reward for the manager policy is excluded due to the performance drop observed in evaluations as in Section \ref{Exp}, and exploration is introduced through the greedy algorithm at the worker level. By doing so, exploration can be performed around the demonstrations. Reward computation during training utilizes sub-rollouts temporally abstracted by selecting every $K$-th step from imagined rollouts. The manager's actor and critics are trained following Director:
\begin{equation}
    \begin{aligned}
        \mathcal{L}(\pi) &= \mathrm{E}[\sum\nolimits_{t=1}^{H-1}\ln\pi(a_t|s_t)\mathrm{sg}(V_t^\lambda-v(s_t))
        +\eta\mathrm{H}[\pi(a_t|s_t)]]\\
        \mathcal{L}(v) &= \mathrm{E}[\sum\nolimits_{t=1}^{H-1}\tfrac{1}{2}(v(s_t)-\mathrm{sg}(V_t^\lambda))^2]
    \end{aligned}
    \label{eq:mgractcrt}
\end{equation}

\subsubsection{Low-level worker policy}

The low-level policy worker generates actions to achieve the subgoal predicted by the manager. Worker is also goal-conditioned by inputting both the current state and the predicted subgoal. It generates action vectors which are then filtered by the selector’s mask, guaranteeing that only actions corresponding to the chosen embodiment are sent to the robot. Formally, the worker policy can be represented as

\begin{equation}
    \begin{aligned}
        \text{Worker Policy:} & & \mathrm{wkr_\psi}(a_t|s_t,sg_t,m_t)
    \end{aligned}
\end{equation}

The worker needs to learn from three rewards: task reward, collision signal, and goal reward. The goal reward measures the similarity between the current state and the subgoal. We implement a greedy algorithm for exploration. Training objectives use Eq.(\ref{eq:mgractcrt}).

\begin{figure*}[!t]
    \centering
    \includegraphics[width=\linewidth]{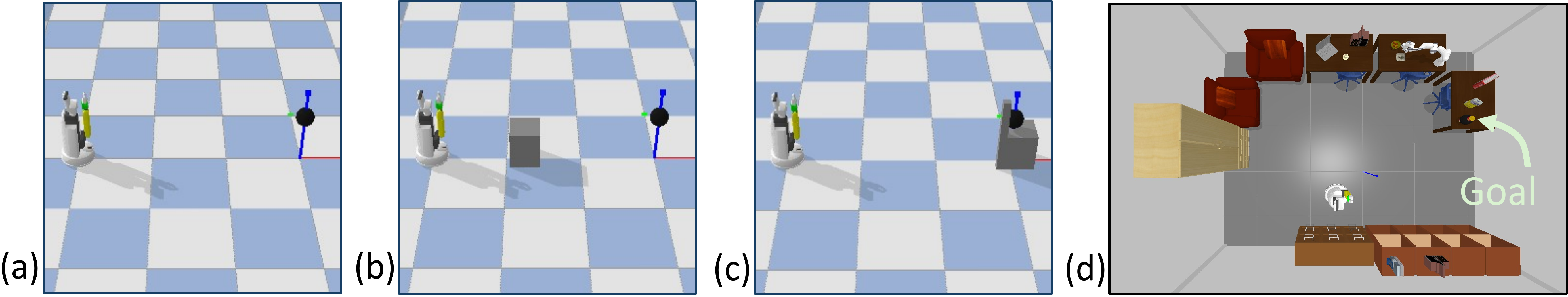}
    \vspace{-0.5cm}
    \caption{Evaluation environments. We prepare four environments to evaluate models. In all environments, the robot is to reach its end-effector to the goal marked by a black sphere. (a) \textbf{Empty:} No obstacle and no collision are considered. (b) \textbf{Obstacle for base:} A fixed block is placed between the robot and the goal, which impedes base motions but has no effect on arm motions near the goal. (c) \textbf{Obstacle for arm:} A block is placed below the goal. A stick is placed diagonally in front of the goal, hindering arm motions from one side. (d) \textbf{Realistic room:} A room space. The goal is located at the corner of one table.}
    \label{fig:env}
    \vspace{-0.3cm}
\end{figure*}

\section{Experiments} \label{Exp}
We evaluate MMDirector through MM reaching tasks in different environments to figure out the following three questions: 
(1) How does MMDirector’s performance compare with those of existing model-based methods?
(2) How does the selector policy influence the embodiment selection?
(3) How does the inclusion of demonstrations impact learning efficiency?
\subsection{Environment Setup}
We conduct reaching tasks using the Toyota HSR robot in Pybullet simulation environments. 
The objective is to navigate the robot to a fixed goal represented by a black sphere. The robot receives RGB and depth images, as well as proprioceptive observations. 
Additionally, 2D ray observations are used when obstacles are present. 
We allow only forward-backward translation and yaw rotation for the base and movements of 3 arm joints. 
The robot's hand remains inactive throughout the experiments. 
We predefined a set of action arrays to represent continuous movements. 
These arrays are indexed based on the one-hot output of the worker policy. 
Then, they are executed in the environment, recorded in the buffer, and subsequently used as input for the world model. 
The initial base position and yaw are sampled randomly within predefined ranges, and success is achieved if the distance between the goal and the midpoint of the two fingertips is less than 0.1m. 
We prepare four environments, as shown in Fig.\ref{fig:env}, and collect demonstrations for each environment.
\begin{itemize}
    \item \textbf{Empty:} We focus on the learning performance of models and the effectiveness of predictive reachability.
    \item \textbf{Obstacle for base:} We place a block between the robot and the goal, and evaluate its impact on base navigation.
    \item \textbf{Obstacles for arm:} We place a block below the goal and a stick diagonally in front of the goal. We focus on the embodiment selection based on predictive reachability when the stick hinders arm motions from one side.
    \item \textbf{Realistic room:} We focus on the embodiment selection based on predictive reachability in a visually complex environment.
\end{itemize}

\subsection{Method Implementation}
We evaluate our proposed MMDirector in the above 4 environments. We also implement the following models for comparison:
\begin{itemize}
    \item \textbf{Director.} Including the world model and two-level policy. 
    \item \textbf{DreamerV2.} Including the world model and flat policy.
    \item \textbf{Demo-as-Experience (DE).} Following MMDirector’s architecture and using demonstrations as experience.
    \item \textbf{Demo-as-Goal (DG).} Following MMDirector’s architecture and using demonstrations as policy goals.
    \item \textbf{NO-demo-for-policy (NO).} Following MMDirector’s architecture and using no demonstration for policy.
\end{itemize}

MMDirector, Director, and DreamerV2 leverage demonstrations as both replay experiences and policy goals, while DE, DG, NO use demonstrations differently. We modified the reward functions accordingly. Director and DG use the same progress reward as MMDirector in their manager policies. DE and NO use an exploration reward for their manager policies instead, as they do not incorporate demonstrations as policy goals. Since DreamerV2 has a flat policy for action generation, its goal reward is computed between the current state and the final goal. Only MMDirector, Director, and DreamerV2 are evaluated in the realistic room.

In pre-training, models are optimized using only demonstrations. In RL phase, the world model is optimized using both demonstrations and online samples, while policies are trained from imagined rollouts. Moreover, the model operates in 3 modes. It predicts rollouts in the imagination mode. In the training sample mode, it collects fixed-length trajectories in the environment and stores them in the replay buffer. During both imagination mode and training sample mode, the $\epsilon$-greedy algorithm is implemented for the selector and worker to introduce exploration into training datasets. In evaluation sample mode, it collects trajectories terminated when goal-reaching or collision occurs. 

The straight-through gradient estimation is performed in policy training. Each training run uses a single V100 or A100 GPU with XLA and mixed precision enabled. Each model is evaluated in 3 different seeds.

\begin{figure*}[!t]
    \centering
    \includegraphics[width=\linewidth]{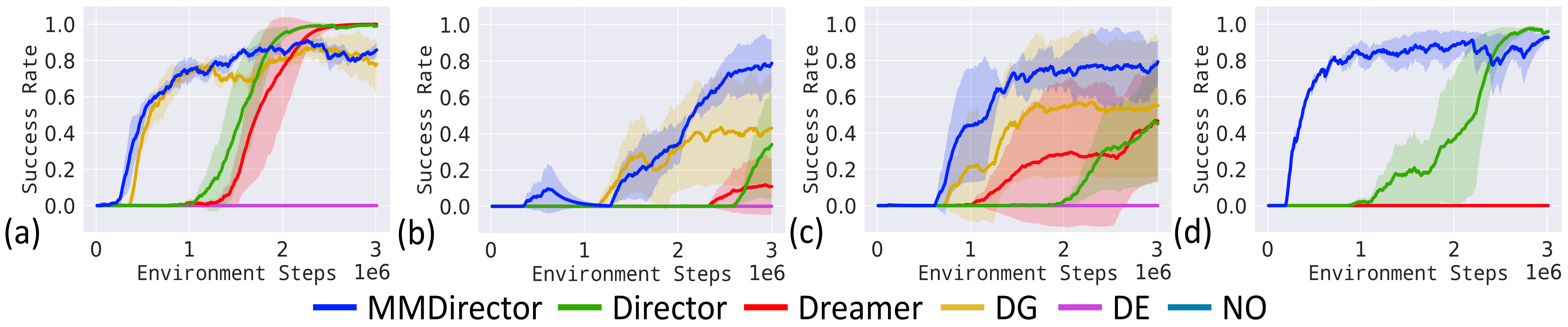}
    \vspace{-0.5cm}
    \caption{Success curves of models in four environments. Only MMDirector, Director, and DreamerV2 are evaluated in the realistic room. MMDirector achieves high success rates across all four environments, though it performs slightly worse than Director and Dreamer in the empty environment and realistic room. The results from DG highlight the effectiveness of using demonstrations as goals for policies.}
    \label{fig:score}
\end{figure*}
\begin{figure*}[!t]
    \centering
    \includegraphics[width=\linewidth]{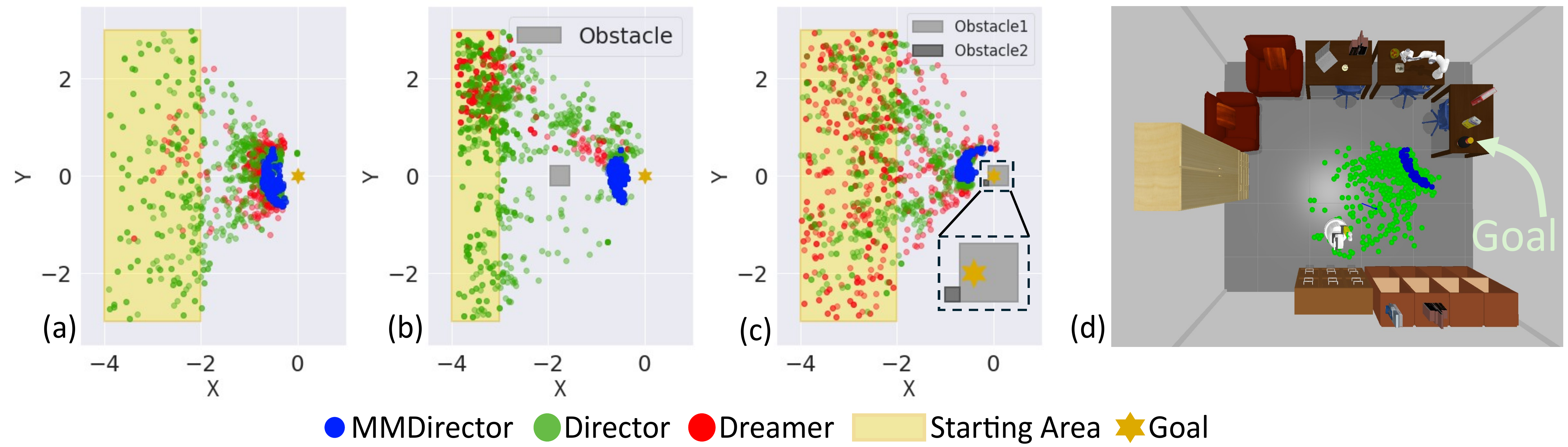}
    \vspace{-0.5cm}
    \caption{Arm selections based on predictive reachability. We visualize the base placements where the arm is selected to move. MMDirector selects arm actions only after the base has approached the goal, whereas Director and Dreamer make arm selections from a distance and even near obstacles.}
    \label{fig:mask}
    \vspace{-3mm}
\end{figure*}

\subsection{Evaluation Results}
\subsubsection{Empty}
We evaluate models by the success rate and embodiment selection in this task. 
Firstly, we compare MMDirector against Director and Dreamer. 
The success rate curves in Fig.\ref{fig:score}(a) show MMDirector’s competitive performance for successful reaching and notable learning acceleration. 
We attribute the speed advantage to exploration within small and separate action spaces rather than the unified one in Director and Dreamer. 
However, MMDirector displays a notable performance drop compared to Director and Dreamer. We analyze 56 failures after 2M steps to understand the causes. 36 of them are due to the inability to generate appropriate arm actions despite correct arm selections. In 15 cases, the base could not reach a suitable position, and the remaining 5 failures result from incorrect arm selections. This suggests that the primary reason for failure is improper policy training for planning. 
\begin{figure}[!t]
    \centering
    \includegraphics[width=\linewidth]{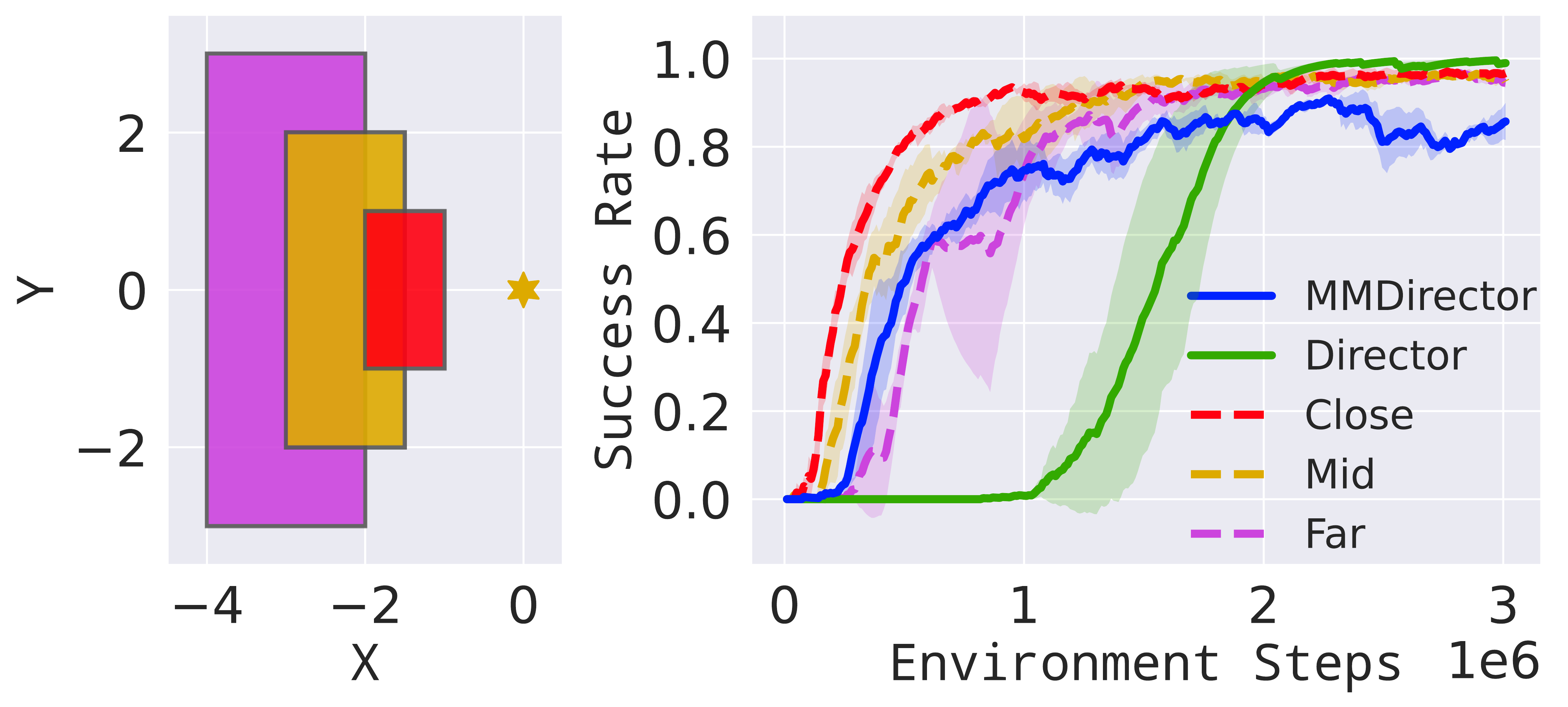}
    \caption{Results of the modified MMDirector with 3 starting areas. The far area is the same as that of the original experiment. While the modified MMDirector shows notable improvements, its performance is still inferior to 100\% of Director.}
    \label{fig:improved}
    \vspace{-5mm}
\end{figure}
We consider two factors accounting for this issue. One is the interdependence between the stages of base and arm. The base stage relies on arm actions to receive task rewards, while the arm policy can only be trained once the base approaches the goal. This interdependence hinders simultaneous and effective training of both policies. The other is the non-stationary issue of HRL optimization caused by varying embodiment selection. To tackle these issues, we make several modifications to MMDirector: (1) it dynamically selects the manager’s stage goal based on the embodiment selection to provide more accurate progress rewards during the base stage, (2) we introduce imitation loss, supervised by the arm actions of demonstrations, to improve the success rate of arm motions, and (3) we completely separate the network architectures for the base and arm at both the manager and worker levels to minimize the impact of HRL-related issues.

The results in Fig.\ref{fig:improved} of testing the modified model across three different starting areas show improved success rates compared with the original. However, the model still poses the same failures as before in all three conditions, leading to suboptimal convergence. We believe that the inherent interdependence remains unresolved. New training strategies or reward designs may be useful to enhance policy effectiveness, which we will explore in future research.

We also analyze the embodiment selection by visualizing base placements when the arm is selected in the successful trajectories. 
The selection is determined by the output mask of the selector in MMDirector and post-processing classification for Director and Dreamer. 
As shown in Fig.\ref{fig:mask}(a), MMDirector selects arm motion only when the base has approached the goal. In contrast, Director and Dreamer make incorrect arm selections at greater distances. 
This highlights the advantage of explicit embodiment selection based on predictive reachability. 

Next, we explore the role of demonstrations as experience and goals. 
The success rate curves of DE and NO reveal the challenges in learning MM behaviors through exploration based on sparse goal-reaching task rewards, even though demonstrations provide successful experiences.
Compared to DE, DG shows that goal-conditioning is critical in this vision-based MM reaching task due to providing dense progress signals in addition to sparse task rewards. 
\begin{figure}[!t]
    \centering
    \includegraphics[width=\linewidth]{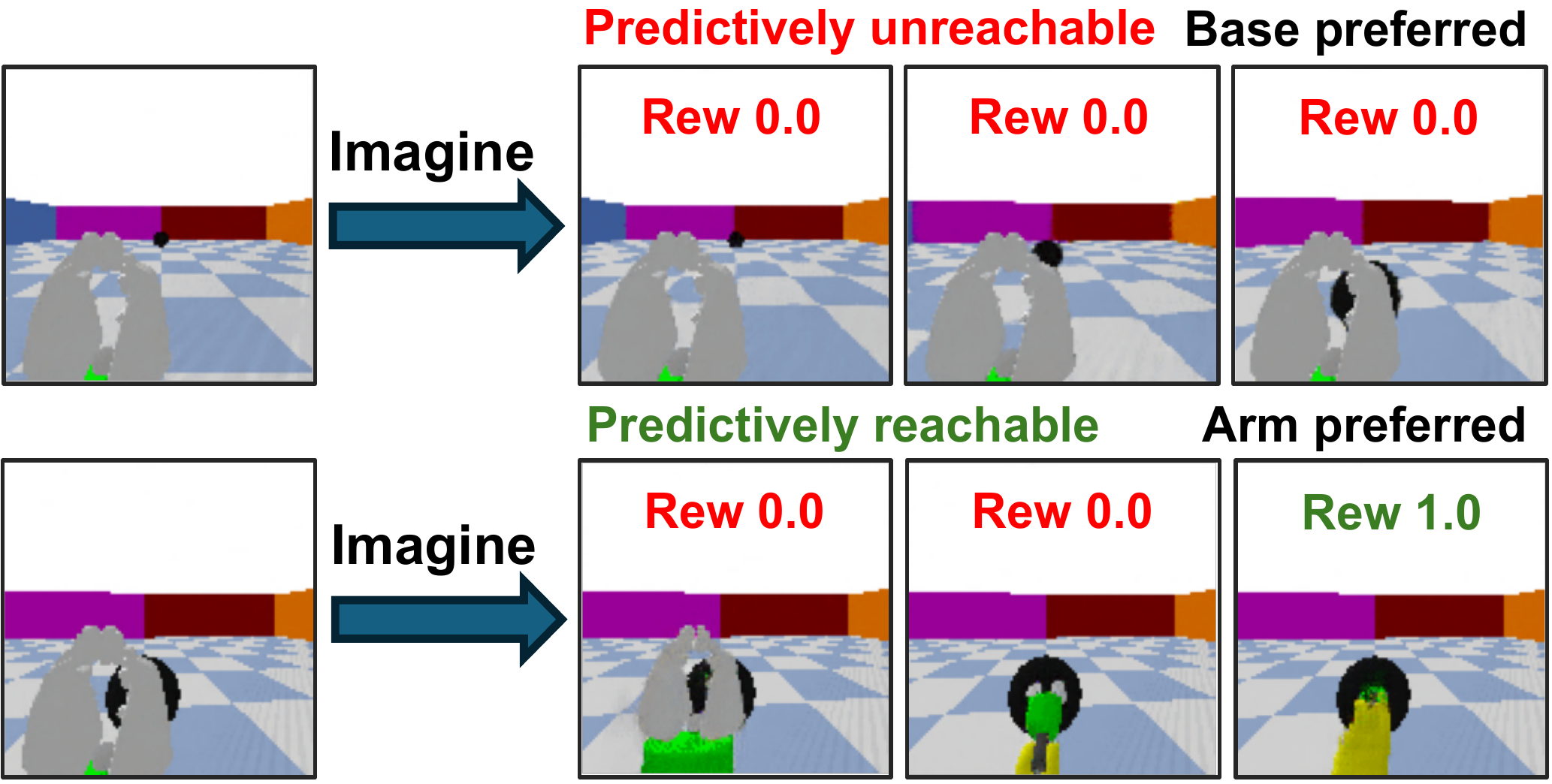}
    \caption{Illustrations of embodiment selections based on predictive reachability. The base is preferred when the goal is predictively unreachable as the task rewards are less than the threshold, while the arm is selected when the goal is reachable as at least one task reward exceeds the threshold. Collision is not considered in the empty environment.}
    \label{fig:PRE}
    \vspace{-5mm}
\end{figure}

Fig.\ref{fig:PRE} illustrates the embodiment selection based on predictive reachability in experiments by visualizing the decoded images and their corresponding predicted task rewards. 
The preferred embodiment for one state is determined based on the predictive reachability inferred from task rewards, following the rules in Fig.\ref{fig:PRI}. Collision is not considered in this empty environment.

\begin{table*}[!t]
    \centering
    \caption{Quantitative evaluation of the selection of arm motion}
    \label{tab:quan}
    \begin{tabular}{c|cccc|cccc}
        \toprule
        \multirow{2}{*}{Model} & \multicolumn{4}{c|}{Average proportion of arm selection near the goal} & \multicolumn{4}{c}{Average ratio of first arm selection to trajectory length} \\ \cmidrule{2-9}
         & Empty & Obstacle for base & Obstacle for arm & Room & Empty & Obstacle for base & Obstacle for arm & Room \\
         \midrule
         MMDirector & \textbf{1.0} & \textbf{1.0} & \textbf{1.0} & \textbf{1.0} & 0.843 (300) & 0.864 (300) & 0.843 (300) & 0.655 (300)\\
         Director & 0.485 & 0.120 & 0.326 & 0.373 & 0.518 (300) & 0.142 (200)  & 0.372 (200) & 0.383 (300)\\
         Dreamer & 0.696 & 0.044 & 0.248 & None & 0.802 (300) & 0.158 (100) & 0.146 (200) & None\\
         \bottomrule
    \end{tabular}
    \vspace{-0.3cm}
\end{table*}

\subsubsection{Obstacle for base}
In this task, we consider an obstacle that impedes base motions. We introduce imitation loss into RL for the manager policy to improve learning. This modification is applied to MMDirector, Director, and DE, as these models utilize both demonstration experiences and the manager policy. The four models using MMDirector’s architecture set the collision reward coefficient to -0.1. The results in Fig.\ref{fig:score}(b) confirm MMDirector's high success rate. However, Director and Dreamer, with a -0.1 collision coefficient, struggle to navigate around the obstacle, making it difficult to explore the environment effectively. We adjust the collision coefficient for Director and Dreamer to -0.01 to relax the collision constraints. The results are shown in Fig.\ref{fig:score}(b). They can achieve more effective exploration and goal-reaching. The results of arm selection in Fig.\ref{fig:mask}(b) are consistent with those in the empty environment. MMDirector still selects the arm near the goal, while Dreamer and Director show instances of moving their arms around the obstacle unnecessarily.

\subsubsection{Obstacle for arm}
We evaluate models in the environment where an obstacle is (obstacle 2) placed diagonally in front of the goal, blocking arm motions on that side. 
The success rate results in Fig.\ref{fig:score}(c) show that MMDirector outperforms all the other models. 
The base placement of MMDirector in Fig.\ref{fig:mask}(c) shows a noticeable adaptation compared to that in the empty environment. 
This indicates that predicted collisions properly filter predictive reachability to provide accurate guidance for arm selections.
By contrast, Director and Dreamer have arm selections on the side of obstacle 2 and away from the goal. 
% This comparison further proves the essential role of explicit embodiment selection based on predictive reachability.

\subsubsection{Realistic room}
We evaluate models in a room space. 
The robot starts from an area and reaches the goal above a table.
Fig.\ref{fig:score}(d) shows that MMDirector maintains the advantages in learning speed and high success rate in reaching the goal. 
However, MMDirector still faces the same policy training issue observed in the empty environment when compared with Director.
Fig.\ref{fig:mask}(d) demonstrates that predictive reachability can effectively guide arm selection at appropriate base positions near the goal when the world model perceives the complex environment.

\subsubsection{Quantitative evaluations}
We provide quantitative assessments of arm selection in Table \ref{tab:quan}. 
Firstly, we quantitatively define that an arm selection is considered near the goal if the 2D distance between the base and the goal is less than 0.8m. 
This value is chosen based on the observation that IK solutions for successful arm reaching exist when the base is located within this range. 
The proportion of arm selections near the goal to the total number of arm selections is then calculated for each model. 
The results demonstrate that MMDirector exhibits the most favorable selections, while Director and Dreamer make incorrect selections beyond the defined range. These are supported by the embodiment selection visualizations in Fig.\ref{fig:mask}. 
Furthermore, we analyze the temporal positioning of arm selection within the trajectory by calculating the ratio of the first arm selection index to the trajectory length. 
This ratio indicates the model’s ability to produce MM behaviors where the arm is activated after the base approaches the goal. 
Numbers in brackets count the trajectories for averaging. 
The results suggest that MMDirector can consistently constrain arm motions near the goal after the base approaches the goal.
However, Director and Dreamer incorrectly select arm motions at positions where the robot base still needs to get closer to the goal.

\subsection{Ablation Studies}
We perform two ablation experiments in the empty environment to verify the role of rewards and the hierarchy for planning.

\begin{figure}[!t]
    \centering
    \includegraphics[width=\linewidth]{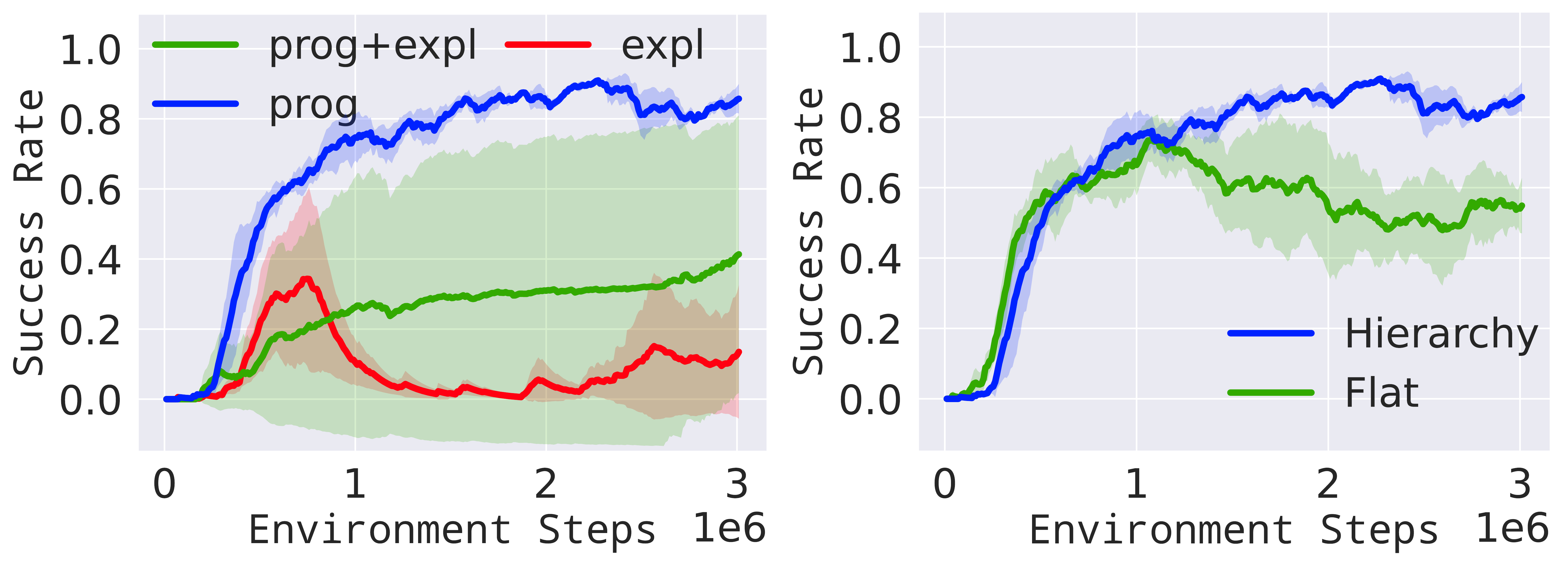}
    \caption{Ablation results. Left: Comparison between progress reward and exploration. Right: Results of using hierarchical and flat policies for planning.}
    \label{fig:abla}
    \vspace{-0.5cm}
\end{figure}
\textbf{Progress reward and exploration reward.} 
To generate effective subgoals, the manager in MMDirector learns from a dense progress reward guided by the final goals in demonstrations, rather than the exploration reward used in the original Director. Though we apply the greedy algorithm in the worker policy for exploration, the exploration at the manager level should be investigated. Here, we compare the performance of using progress rewards versus exploration rewards. As shown in Fig.\ref{fig:abla}(Left), the progress reward effectively guides the robot to the goal, but the exploration reward results in poor performance. This suggests that exploration based solely on image observations and sparse task rewards is inefficient and highlights the importance of demonstrations in guiding the learning process.

\textbf{Hierarchy.} 
In MMDirector, we employ a hierarchy consisting of the manager and worker for planning. To evaluate the impact of hierarchical planning, we compare it with a new model, which removes the manager, leaving only the flat worker to handle planning. The final goal $stg_2$ is used as the policy goal. As shown in Fig.\ref{fig:abla}(Right), the hierarchical policy yields better and more stable success rates over the flat one. The main issue of the flat lies in the base's inability to approach the goal. This indicates that for learning the base policy, utilizing intermediate subgoals offers more effective guidance compared to directly using the distant final goal.

\section{Conclusion}
This work introduces predictive reachability to facilitate embodiment selection in the coordinated MM behaviors in vision-based tasks. 
We propose MMDirector, which learns embodiment selection based on predictive reachability and plans actions accordingly. The separation of the action spaces of the base and arm simplifies exploration and accelerates learning. Experiments show that MMDirector achieves better learning performance and properly selects the arm near the goal and the base at a distance. Additionally, demonstrations significantly improve the performance by introducing the dense progress reward.

However, our model has limitations. The interdependence between the base and arm stages hinders simultaneous and effective training of both planning policies. Compared to the reachability w.r.t. the system, the predictive reachability w.r.t. a learned policy may cause unstable reachability estimations and further incorrect embodiment selections. Furthermore, this work only focuses on simple reaching tasks. Applying predictive reachability in complex tasks, such as grasping, and in real-world scenarios requires further investigation. We aim to address these limitations in future studies.

\section*{Acknowledgment}
This work was supported by the JST Moonshot R\&D Grant Number JPMJMS2011.

% \begin{thebibliography}{1}
\bibliographystyle{IEEEtran}
\bibliography{./bibtex/bib/paper}

\end{document}